\pdfoutput=1

\documentclass[11pt]{article}

\usepackage[]{acl}

\usepackage{times}
\usepackage{latexsym}
\usepackage[prependcaption,textsize=tiny]{todonotes}

\usepackage[T1]{fontenc}

\usepackage[utf8]{inputenc}
\usepackage{adjustbox}
\usepackage{multirow}
\usepackage[normalem]{ulem}
\usepackage{wrapfig}
 \usepackage{booktabs}
 \usepackage{subfig}
\usepackage{microtype}
\usepackage{soul}

\title{Prabhupadavani: A Code-mixed Speech Translation Data for 25 Languages}

\newcommand*{\affmark}[1][*]{\textsuperscript{#1}}

\author{Jivnesh Sandhan\affmark[1], Ayush Daksh\affmark[2], Om Adideva Paranjay\affmark[3],\\\textbf{Laxmidhar Behera\affmark[1,4] and Pawan Goyal\affmark[2]}\\
\affmark[1]IIT Kanpur, \affmark[2]IIT Kharagpur, \affmark[3]University of Pennsylvania, \affmark[4]IIT Mandi\\
\texttt{jivnesh@iitk.ac.in, pawang@cse.iitkgp.ac.in}}
\renewcommand\footnotemark{}

\begin{document}
\maketitle
\begin{abstract}
Nowadays, the interest in code-mixing has become ubiquitous in Natural Language Processing (NLP); however, not much attention has been given to address this phenomenon for Speech Translation (ST) task. This can be solely attributed to the lack of code-mixed ST task labelled data. Thus, we introduce Prabhupadavani, which is a multilingual code-mixed ST dataset for 25 languages. It is multi-domain, covers ten language families, containing 94 hours of speech by 130+ speakers, manually aligned with corresponding text in the target language. The Prabhupadavani is about Vedic culture and heritage from Indic literature, where code-switching in the case of quotation from literature is important in the context of humanities teaching. To the best of our knowledge, Prabhupadvani is the first multi-lingual code-mixed ST dataset available in the ST literature. This data also can be used for a code-mixed machine translation task. All the dataset can be accessed at:
\url{https://github.com/frozentoad9/CMST}.
\end{abstract}

\section{Introduction}
\label{intro}
Speech Translation (ST) is a task in which speech is simultaneously translated from source language to a different target language.\footnote{We refer to speech translation as a speech-to-text task.} It aids to overcome the language barriers across different communities for various applications such as social media, education, tourism, medical etc. 
Earlier attempts to build a robust speech translation system mainly focused on a cascaded approach where two separate architectures for Automatic Speech Recognition (ASR) and Machine Translation (MT) are used in pipeline mode \cite{cho2013real,post2013improved,tsvetkov-etal-2014-augmenting,ruiz2015adapting,sperber2017toward}.
 However, these approaches mainly suffer from cascading effect of error propagation.
Thus, attention shifted to end-to-end approaches \cite{berard2018end,duong-etal-2016-attentional,weiss2017sequence,bansal-etal-2017-towards} due to their ability to obliviate error propagation and ease of maintaining a single architecture. However, these end-to-end approaches could not match the performance of cascaded systems due to the lack of sufficiently large data \cite{niehues-etal-2021-tutorial}. Notably, with the recent upsurge in ST datasets \cite{di-gangi-etal-2019-must,zanon-boito-etal-2020-mass,jairsan2020a,wang-etal-2020-covost}, this gap has been closed \cite{niehues-etal-2021-tutorial,ansari2020findings}.

Nowadays, most users prefer to communicate using a mixture of two or many languages on platforms such as social media, online blogs, chatbots, etc. Thus, code-mixing has become ubiquitous in all kinds of Natural Language Processing (NLP) resources/tasks \cite{khanuja2020new,chakravarthi2020corpus,singh2018twitter,singh2018named,dhar2018enabling}.  However, the existing NLP tools may not be robust enough to address this phenomenon of code-mixing for various downstream NLP applications \cite{srivastava2021challenges}. Therefore, there has been a surge in creating code-mixed datasets: (1) to understand reasons for the failure of existing models, and (2) to empower existing models for overcoming this phenomenon. Nevertheless, it is challenging to find natural resources that essentially capture different aspects of code-mixing for creating datasets for a wide range of NLP tasks.
Although there has been considerable research in generating copious data and novel architectures for the ST task, we find that not much attention has been given to address the code-mixing phenomenon on the ST task. Possibly, this can be justified due to the lack of a code-mixed ST dataset.
To the best of our knowledge, no such sufficiently large, multi-lingual, naturally occurring, code-mixed dataset is available for the ST. 

Thus, in this work, we introduce \textbf{Prabhupadavani}, a multi-lingual, multi-domain, speech translation dataset for 25 languages containing 94 hours of speech by 130 speakers. The Prabhupadavani is about Vedic culture and heritage from Indic literature, where code-switching in the case of quotation from literature is important in the context of humanities teaching. The multiple domains cover utterances from public lectures, conversations, debates, and interviews on various social issues.
This is the first code-mixed data for speech translation to the best of our knowledge. It is code-mixed with English, Bengali and Sanskrit. From the typological point of view, the languages covered vary over ten language families. All the audios files have been manually aligned and translated. 
 We believe that our work will ignite research in this direction to understand- (1) How to make existing systems robust for handling this phenomenon effectively? (2) Can multi-lingual training help to improve performance on code-mixed speech translation? (3) Will the gap between the cascade and end-to-end systems be closed? (4) Can we train a single model for all languages using parallel nature of the dataset?\footnote{Prabhupadavani has parallel translations available in all the 25 languages for all the utterances.}

\begin{table*}[h]
\centering
\begin{small}
\resizebox{0.7\textwidth}{!}{%
\begin{tabular}{l|rcccc}
\hline
\textbf{Languages} & \multicolumn{1}{l}{\textbf{\# Types}} & \multicolumn{1}{l}{\textbf{\# Tokens}} & \multicolumn{1}{l}{\textbf{Types per line}} & \multicolumn{1}{l}{\textbf{Tokens per line}} & \multicolumn{1}{l}{\textbf{Avg. token length}} \\
\hline
English            & 40,324                               & 601,889                               & 10.58                                       & 11.27                                        & 4.92                                           \\
French (France)    & 50,510                               & 645,651                               & 11.38                                       & 12.09                                        & 5.08                                           \\
German             & 50,748                               & 584,575                               & 10.44                                       & 10.95                                        & 5.57                                           \\
Gujarati           & 41,959                               & 584,989                               & 10.37                                       & 10.95                                        & 4.46                                           \\
Hindi              & 29,744                               & 716,800                               & 12.36                                       & 13.42                                        & 3.74                                           \\
Hungarian          & 84,872                               & 506,608                               & 9.13                                        & 9.49                                         & 5.89                                           \\
Indonesian         & 39,365                               & 653,374                               & 11.54                                       & 12.23                                        & 6.14                                           \\
Italian            & 52,372                               & 512,061                               & 9.23                                        & 9.59                                         & 5.37                                           \\
Latvian            & 70,040                               & 477,106                               & 8.69                                        & 8.93                                         & 5.72                                           \\
Lithuanian         & 75,222                               & 491,558                               & 8.92                                        & 9.20                                         & 6.04                                           \\
Nepali             & 52,630                               & 570,268                               & 10.03                                       & 10.68                                        & 4.88                                           \\
Persian (Farsi)    & 51,722                               & 598,096                               & 10.61                                       & 11.20                                        & 4.10                                           \\
Polish             & 71,662                               & 494,263                               & 8.99                                        & 9.25                                         & 5.86                                           \\
Portuguese(Brazil) & 50,087                               & 608,432                               & 10.80                                       & 11.39                                        & 5.12                                           \\
Russian            & 72,162                               & 490,908                               & 8.96                                        & 9.19                                         & 5.79                                           \\
Slovak             & 73,789                               & 520,465                               & 9.39                                        & 9.75                                         & 5.37                                           \\
Slovenian          & 68,619                               & 516,649                               & 9.35                                        & 9.67                                         & 5.30                                           \\
Spanish            & 49,806                               & 608,868                               & 10.75                                       & 11.40                                        & 5.07                                           \\
Swedish            & 48,233                               & 581,751                               & 10.31                                       & 10.89                                        & 5.00                                           \\
Tamil              & 84,183                               & 460,678                               & 8.37                                        & 8.63                                         & 7.65                                           \\
Telugu             & 72,006                               & 464,665                               & 8.34                                        & 8.70                                         & 6.56                                           \\
Turkish            & 78,957                               & 453,521                               & 8.27                                        & 8.49                                         & 6.35                                           \\
Bulgarian          & 60,712                               & 564,150                               & 10.10                                       & 10.56                                        & 5.24                                           \\
Croatian           & 73,075                               & 531,326                               & 9.58                                        & 9.95                                         & 5.28                                           \\
Danish             & 50,170                               & 587,253                               & 10.40                                       & 11.00                                        & 4.98                                           \\
Dutch              & 42,716                               & 595,464                               & 10.52                                       & 11.15                                        & 5.05      
                            \\
\hline
\end{tabular}}
 \end{small}
\caption{Statistics of the \textbf{Prabhupadavani} dataset}
\end{table*}
\section{Related Work}
\label{related_work}

\paragraph{Speech Translation:} 
Recently, there have been increased efforts for creating large speech translation data sets for many languages. Nevertheless, the available datasets are limited to certain languages or only underpaid licenses for non-English languages. Most of the relatively larger datasets are English-centric \cite{di-gangi-etal-2019-must,berard2018end}, domain-specific \cite{zanon-boito-etal-2020-mass,jairsan2020a} or with limited speech hours \cite{zanon-boito-etal-2020-mass,zhang2021bstc}.
Recently, there has been upsurge in ST datasets in the literature \cite{di-gangi-etal-2019-must,jairsan2020a,wang-etal-2020-covost,salesky2021mtedx}.
To the best of our knowledge, there is no such naturally occurring, sufficiently large and multi-lingual ST dataset that contains a code-switching phenomenon. 
We fill this gap by contributing a code-mixed speech translation dataset for 25 languages.


\paragraph{Code-mixing:}
  Code-mixing is ubiquitous and well addressed on variety of downstream NLP tasks.
 However, majority of code-mixed datasets are synthetically generated \cite{gonen2018language,khanuja2020new}; therefore, they may not be able to capture the different aspects of code-mixing.
The code-mixed datasets for an ASR task are either limited to only one/two languages or contain only a few hours of speech data \cite{Nakayama,lyu2015mandarin}.
 Synthetic code-mixed datasets may not capture the different aspects of code-mixing. 
To the best of our knowledge, Prabhupadavani is the first code-mixed dataset available for 25 languages on speech translation task.  

\section{Data Description}
\paragraph{Resource:} Vanimedia's Multi-language Subtitle Project\footnote{\url{https://vanimedia.org/wiki/Multi-language_Subtitle_Project}} has created 1,080 audio mini-clips of \'{S}r\={i}la Prabhup\={a}da's lectures, conversations, debates, interviews and is now transcribing them in multiple languages.\footnote{\url{https://vanimedia.org/wiki/Table:_Clips_to_subtitle}}
700+ translators participate in creating subtitles for all 1,080 mini-clips for 108+ languages. Currently, this work has been completed for 25 languages. To procure subtitles for each clip in multiple languages, translators are provided with mini-clips and English subtitles that are manually aligned with each utterance. They use a third-party software named Dotsub.com\footnote{\url{https://dotsub.com}} and the task of translators is to provide translation for the corresponding utterance with the help of given transcription. 
On average, there are 3-4 translators for each language, and each clip takes more or less one hour for translation. Each translator has invested an average of 6 hours every day in translating these clips. Collectively, the time taken to translate 1,080 clips into 25 languages is over 45 weeks. 
Current release of the dataset contains 25 languages (including transcription) for which transcription/translations are available. For these languages, we have over 53K utterances of transcription and translations. 
\begin{table}[t]
\begin{small}
\centering
\begin{tabular}{|c|c|c|c|}
\hline
\textbf{Language} & \textbf{Tokens} & \textbf{Types} & \textbf{Percentage} \\
\hline
English           & 500,136              & 6,312                 & 83.6                \\
Bengali           & 46,933               & 3,907                 & 7.84                \\
Sanskrit          & 51,246               & 7,202                 & 8.56                \\
\hline
\textbf{Total}    & 598,315              & 17,421                & 100                \\
\hline
\end{tabular}
\caption{Statistics of code-mixing in \textbf{Prabhupadavani}}
\label{table:statistics}
\end{small}
\end{table}

\paragraph{Preprocessing:}
In this section, we describe pre-processing steps followed to arrive at the final version of the data.
First, we scrape transcription and their translations in 25 languages from Dotsub and extracted the corresponding Youtube links of all audio clips. We use Selenium\footnote{\url{https://www.selenium.dev/}} web crawler to automatize the process of downloading. We convert those videos into MP3 audio clips using a third-party application\footnote{\url{https://ytmp3.cc/uu100cc/}}. We chop the converted audio files based on the timestamps provided in the subtitle (.srt) files.\footnote{\url{https://pypi.org/project/audioclipextractor/}} This process boils down to 53,000 utterances with their transcription and translation in 25 languages. Table ~\ref{table:source-target-examples} illustrates the example from Prabhupadavani. In order to obliviate time and efforts needed for data pre-processing, we provide train, dev and test set splits using stratified sampling. There are 51,000, 1,000 and 1,000 utterances in train, dev and test set, respectively. We consider the following dimensions for stratified sampling: (1) different speakers (2) proportion of intra-sentential and inter-sentential code-mixing.
\begin{table*}[t]
\begin{small}
\centering
\begin{tabular}{|c|c|c|}
\hline
\textbf{Language} &\textbf{Type} & \textbf{Examples}\\
\hline
English-Sanskrit& Inter & K\d{r}\d{s}\d{n}a is assuring. \textcolor{red}{aha\.{m} tv\={a}\.{m} sarva-p\={a}pebhyo mok\d{s}ayi\d{s}y\={a}mi} \\
&Intra&Sense gratification means \textcolor{red}{udara-upastha-jihv\={a}}\\
\hline
Sanskrit-English&Inter&\textcolor{red}{\={I}\`{s}\={a}v\={a}syam ida\.{m} sarvam}. Everything belongs to God \\
&Intra&\textcolor{red}{andh\={a} yath\={a}ndhair upan\={i}yam\={a}n\={a}\d{h}} and people, leaders. \\ \hline
English-Bengali&Inter&Give up their. \textcolor{violet}{Asat-sa\.{n}ga-ty\={a}ga ei vai\d{s}\d{n}ava \={a}c\={a}ra}\\
&Intra& Therefore Caitanya Mah\={a}prabhu said \textcolor{violet}{guru-k\d{r}\d{s}\d{n}a-k\d{r}p\={a}ya}\\ \hline
Bengali-English&Inter&\textcolor{violet}{Guru-k\d{r}\d{s}\d{n}a-k\d{r}p\={a}ya p\={a}ya bhakti-lat\={a}-b\={i}ja}. Then our devotional service is perfect\\
&Intra&\textcolor{violet}{t\={a}\.{n}h\={a}ra n\={a}hika do\d{s}a} means he is not faulty.\\\hline
\end{tabular}
\caption{Examples of code-mixing in \textbf{Prabhupadavani}. English, Sanskrit and Bengali are indicated by black, red and violet color, respectively.}
\label{table:code-mixed_examples}
\end{small}
\end{table*}
\begin{table*}[t]
\begin{small}
\centering
\begin{tabular}{|c|c|c|}
\hline
\textbf{} &\textbf{Language} & \textbf{Translations}\\
\hline
Source& English &You can become Brahman. Brahma-bh\={u}y\={a}ya kalpate  \\ \hline
Target& Bulgarian &\includegraphics[width=0.45\textwidth]{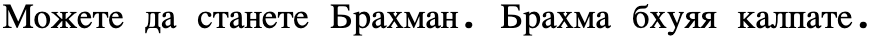}\\
& Hindi &\includegraphics[width=0.4\textwidth, height=0.35cm]{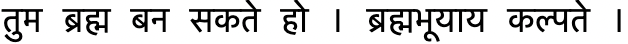}\\
& Russian &\includegraphics[width=0.45\textwidth]{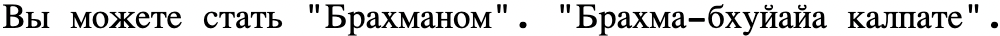}\\
& Tamil &\includegraphics[width=0.45\textwidth]{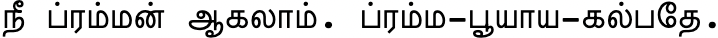}\\
& Gujarati &\includegraphics[width=0.45\textwidth, height=0.35cm]{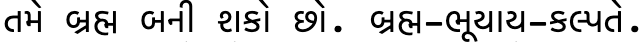}\\
\hline
\end{tabular}
\caption{Sample data point from \textbf{Prabhupadavani}. For a code-mixed utterance, we show its English transcription (source) and the corresponding translations for 5 languages.}
\label{table:source-target-examples}
\end{small}
\end{table*}

\paragraph{Code-mixing:} 
Prabhupadavani is code-mixed across three languages: English, Sanskrit, and Bengali. Table~\ref{table:statistics} reports the overall statistics of the code-mixing present in Prabhupadavani. If we consider the number of tokens in utterances, then it is mainly dominated by English (83.0\%) tokens; however, it is not the case in terms of a number of types. This attributes to the contrasting nature of morphology (English vs Sanskrit/Bengali).
Code-mixing is categorized into two classes- (1) inter-sentential: speaker chooses to switch the language after completion of utterance (2) intra-sentential: speaker switches the language within an utterance. Table \ref{table:code-mixed_examples} illustrates examples of code-mixing from Prabhupadavani. Table \ref{table:ablation_train_size} shows the statistics of both these types of code-mixing.
Mainly speaker explains Sanskrit verses to the English audience; therefore, transitions between the Sanskrit-English pair is more. However, sometimes speaker also use Bengali literature to illustrate the points. Thus, we observe Bengali-English code-switching. Notably, there is no code-switching between Bengali-Sanskrit because the audience is English speaking.

 \begin{table}[h]
  \begin{small}
\centering
\begin{tabular}{|c|c|c|}
\hline
     &\textbf{Inter-Sentential} & \textbf{Intra-Sentential}           \\
\hline
English-Sanskrit  & 2,356 & 2,338  \\
Sanskrit-English  & 2366 & 851 \\
English-Bengali  & 339 & 124 \\
Bengali-English  & 339 & 0 \\
\hline
\end{tabular}
    \caption{Code-Mixing type for our dataset}
     \label{table:ablation_train_size}
     \end{small}
\end{table}

\paragraph{Language diversity:} From typological point of view, Prabhupadavani covers 25 languages (inlcluding ASR) from 10 language families. They are listed as follows:
(1) \textit{Indo-European:-}
(a) \textit{Romance:}  Italian, Portuguese, Spanish, French
(b) \textit{Germanic:}  German, Swedish, Danish, Dutch, English
(c) \textit{Baltic:} Lithuanian, Latvian 
(d) \textit{Slavic:} Croatian, Polish, Slovak, Russian, Slovenian, Bulgarian, 
(e) \textit{Indo-Aryan:} Bengali, Hindi, Sanskrit, Nepali
(f) \textit{Indo-Iranian:} Persian (Farsi)
(2) \textit{Uralic:-}
(a) \textit{Finno-Ugric:} Hungarian
(3) \textit{Austronesian:-}
(a) \textit{Malayo-Polynesian:} Indonesian
(4) \textit{Dravidian:-}
Tamil, Telugu.
Prabhupadavani contains fusional languages: Indo-European, Uralic and Agglutinative languages: Austronesian (Indonesian), Dravidian and Turkic (Turkish). In the former languages, grammatical markers bear several meanings and for the latter ones, they exhibit only one meaning at the same time.
Except Hindi (Indo-Aryan), all the languages in our dataset use nominative-accusative marking. Hindi uses ergative-absolutive marking upto a limited extent. 
We can also categorize languages based on the number of grammatical genders. Some languages pose (1) three genders: German, Russian, Swedish, etc. (2) two genders: French, Spain, Hindi, etc. (3) no genders: English, Nepali, Persian, Turkish, etc.
Based on a syntactic construct, we can categorize the languages present in Prabhupadavani- (1) SVO word order: English, Italian, French, Indonesian, etc. (2) SOV word order: Indo-Aryan, Dravidian and Turkic (3) flexible word order: Russian, Hungarian.

Thus, the diversified language coverage of \textbf{Prabhupadavani} along with its code-mixed nature makes it a suitable dataset to investigate various linguistic phenomena for the speech translation task, ASR and machine translation. 

\paragraph{Applications of dataset:}
In this section, we throw some light on possible applications of Prabhupadavani: (1) The full dataset of Prabhupadavani contains 2,400 hours of speech. The current release of Prabhupadavani dataset can be utilized to facilitate automatic subtitling of the remaining part of data in the different languages. In this way, it will also be helpful to generate relatively larger dataset.
(2) This dataset can provide a fertile soil to investigate on- How to make existing systems robust to code-mixing phenomenon? Will the gap between cascade and end-to-end to approaches be closed? Can multi-lingual training help to address code-mixing phenomenon? Can we train single ST system for all languages? How robust will be these models on another domain?

\section{Conclusion and Discussion}
\label{conclusion}
In this work, we focused on a code-mixed speech translation dataset. Although code-mixing is a spoken language phenomenon, not much attention has been given to address this phenomenon due to the unavailability of such a dataset. Thus, we released  Prabhupadavani, a high-quality multilingual multi-domain code-mixed ST dataset containing 94 hours of speech data, 130+ speakers, for 25 languages covering 10 language families. The dataset is code-mixed with three languages: English, Bengali, and Sanskrit. In order to reduce the efforts needed for pre-processing, we provide stratified data splits for the dataset. The same dataset can be utilized for the code-mixed machine translation task. 
We believe that these efforts will (1) set a fertile soil for investigating the applicability of existing solutions, (2) help to analyze the kind of errors existing systems are making, and (3) facilitate researchers to propose a novel solution to make existing systems robust for code-mixing. We plan to extend this dataset for 108+ languages. 

In code-switching conversations, speakers prefer a specific communication in a specific language of choice. In that context, the interesting factor is often when the boundary between languages becomes fuzzy, as in cases where an English verb stem is used with Spanish morphology. In the case of Prabhupadavani, the audience does not necessarily speak Sanskrit or Bengali, and the code-switching is primarily quotation or explanation. That is interesting as a phenomenon, but it is not the same as dual bilingual code-switching. We believe future work may help to get deep insights into this phenomenon.


\paragraph{Ethics Statement:}
We do not foresee any ethical concerns with the
work presented in this manuscript. We have taken the consent of the Vanipedia team and Bhaktivedanta Book Trust International to use translations and audio in our dataset.

\section*{Acknowledgements}
We would like to thank the Vanipedia team (\url{https://vanipedia.org/}) of 700+ translators for establishing this multi-lingual database for us to develop. We thank the Bhaktivedanta Book Trust International for permitting us to use Prabhupadavani audio in our dataset. We are grateful to Manish Gupta, Microsoft, for helping us with insightful discussions. We would like to thank the anonymous reviewers for their constructive feedback towards improving this work.  The TCS Fellowship supports the first author's work under Project TCS/EE/2011191P.  

\bibliography{anthology,custom}
\bibliographystyle{acl_natbib}
\appendix

\end{document}